\documentclass[9pt, conference]{IEEEtran}
\IEEEoverridecommandlockouts
\usepackage{cite}
\usepackage{amsmath,amssymb,amsfonts}
\usepackage{algorithmic}
\usepackage{graphicx}
\usepackage{booktabs}
\usepackage{textcomp}
\usepackage{xcolor}
\usepackage[acronym]{glossaries}
\usepackage{multirow}
\usepackage{hyperref}
\usepackage{cleveref}
\usepackage{subfig}
\usepackage{inconsolata}
\usepackage{relsize}
\usepackage{enumitem}
\def\BibTeX{{\rm B\kern-.05em{\sc i\kern-.025em b}\kern-.08em
    T\kern-.1667em\lower.7ex\hbox{E}\kern-.125emX}}

\makeglossaries
\newacronym{aig}{AIG}{And-Inverter Graph}
\newacronym{bo}{BO}{Bayesian Optimization}
\newacronym{cmplx}{\textit{CPX}}{Complexity}
\newacronym{da}{DA}{Data Augmentation}
\newacronym{dse}{DSE}{Design Space Exploration}
\newacronym{eda}{EDA}{Electronic Design Automation}
\newacronym{fsm}{FSM}{Finite State Machines}
\newacronym{iism}{IISM}{Inter-Iteration Selection Module}
\newacronym{llm}{LLM}{Large Language Models}
\newacronym{lso}{LSO}{Logic Synthesis Optimization}
\newacronym{lut}{LUT}{Look-Up-Table}
\newacronym{mig}{MIG}{Majority-Inverter-Gate}
\newacronym{mse}{MSE}{Mean Square Error}
\newacronym{mt}{MT}{Mockturtle}
\newacronym{na}{NA}{Naive Approach}
\newacronym{ni}{NI}{Network Inversion}
\newacronym{ntrans}{\textit{NT}}{Number of Transistors}
\newacronym{prm}{PrM}{Prediction Module}
\newacronym{pom}{PoM}{Policy Module}
\newacronym{pow}{\textit{P}}{Power}
\newacronym{qmp}{QMP}{Quality Metric Predictor}
\newacronym{qor}{QoR}{Quality of Results}
\newacronym{rmse}{RMSE}{Root Mean Square Error}
\newacronym{rtl}{RTL}{Register-Transfer Level}
\newacronym{sa}{SA}{Step-Ahead}
\newacronym{sme}{SME}{Sign-Magnitude Extended}
\newacronym{sm}{SM}{Sign-Magnitude}
\newacronym{sota}{SoTA}{State-of-The-Art}
\newacronym{ssl}{SSL}{Self-Supervised Learning}
\newacronym{swact}{\textit{SwAct}}{Switching Activity}
\newacronym{tc}{TC}{Two's Complement}

\begin{document}

% \title{DENIAL: \textbf{D}esign Space \textbf{E}xploration via \textbf{N}etwork \textbf{I}nversion for Power Efficient Algorithmic Logic Unit for Tensor Processing}
\title{GENIAL: \textbf{G}enerative Design Space \textbf{E}xploration via \textbf{N}etwork \textbf{I}nversion for Low Power \textbf{A}lgorithmic \textbf{L}ogic Units}
% \title{GENIAL: \textbf{G}enerative \textbf{E}ncoding Space Exploration via \textbf{N}etwork \textbf{I}nversion for Low Power \textbf{A}lgorithmic \textbf{L}ogic Units}

% \thanks{}
%     A bit of text.
% }
% Reduce space between authors and text
\IEEEaftertitletext{\vspace{-2\baselineskip}}

% \author{\IEEEauthorblockN{Anonymous Authors}
% \IEEEauthorblockA{Undisclosed Affiliation}}

\author{\IEEEauthorblockN{Maxence Bouvier}
\IEEEauthorblockA{Huawei, Switzerland}
\and
\IEEEauthorblockN{Ryan Amaudruz}
\IEEEauthorblockA{Huawei, Switzerland}
\and
\IEEEauthorblockN{Felix Arnold}
\IEEEauthorblockA{Huawei, Switzerland}
\and
\IEEEauthorblockN{Renzo Andri}
\IEEEauthorblockA{Huawei, Switzerland}
\and
\IEEEauthorblockN{Lukas Cavigelli}
\IEEEauthorblockA{Huawei, Switzerland}
}

\maketitle

\thispagestyle{plain}
\pagestyle{plain}

\begin{abstract}
As AI workloads proliferate, optimizing arithmetic units is becoming increasingly important for reducing the footprint of digital systems.
Conventional design flows, which often rely on manual or heuristic-based optimization, are limited in their ability to thoroughly explore the vast design space.
In this paper, we introduce GENIAL, a machine learning-based framework for the automatic generation and optimization of arithmetic units, with a focus on multipliers.

At the core of GENIAL is a Transformer-based surrogate model trained in two stages, involving self-supervised pretraining followed by supervised finetuning, to robustly forecast key hardware metrics such as power and area from abstracted design representations.
By inverting the surrogate model, GENIAL efficiently searches for new operand encodings that directly minimize power consumption in arithmetic units for specific input data distributions.
Extensive experiments on large datasets demonstrate that GENIAL is consistently more sample efficient than other methods, and converges faster towards optimized designs.
This enables deployment of a high-effort logic synthesis optimization flow in the loop, improving the accuracy of the surrogate model.
Notably, GENIAL automatically discovers encodings that achieve up to 18\% switching activity savings within multipliers on representative AI workloads compared with the conventional two's complement.
We also demonstrate the versatility of our approach by achieving significant improvements on Finite State Machines, highlighting GENIAL's applicability for a wide spectrum of logic functions.
Together, these advances mark a significant step toward automated Quality-of-Results-optimized combinational circuit generation for digital systems.

\end{abstract}


\section{Introduction}
\label{sec:introduction}
% \subsection{Motivation}
% AFTER
This work aims to provide a modern look at how values are represented in binary format, with the ultimate goal of reducing the energy consumption of algorithmic logic units in AI accelerators and associated communication devices.
An encoding is a mapping between a set of values and their corresponding binary representations.
There are numerous well-recognized encodings such as the two's complement and sign-magnitude for fixed-point numbers\cite{arnold2025explicitsignmagnitudeencoders}, the IEEE standard floating-point representations\cite{ieee754}, and the more modern \textit{Brain Float} from Google\cite{intel-bfloat16-2020, wang2019bfloat16} and \textit{HiFloat} from Huawei\cite{luo2024ascendhifloat8formatdeep}.
Most of these encodings were designed to support fast and efficient binary arithmetic operations. However, they were not specifically designed to reduce internal switching activity, which is the main source of power consumption in compute-bound applications.
The broadly implemented encoding for signed integer values, the \gls{tc}, exhibits significant bit-level toggling for operand values close to zero and may therefore be suboptimal in terms of energy efficiency for workloads with value distributions centered around zero.
The \gls{sm} encoding demonstrates less toggling around zero and, as a result, shows reduced switching activity for the multiplication operations\cite{arnold2025explicitsignmagnitudeencoders}.
This raises the question: are there encodings that could decrease switching activity even further than \gls{sm}? 

The proliferation of \gls{llm}-based chatbots and other compute-intensive generative applications is continuously increasing the global energy demand of data centers.
Besides being hosted on cloud platforms, many LLMs are deployed at the edge in phones, laptops, etc.
In both scenarios, quantizing weights and activations to a smaller set of integer values is a common strategy to reduce computational, communication, and memory overheads, enabling more efficient low-precision computation.
Prior studies have shown that 4-bit quantization can have a negligible impact on model accuracy \cite{NEURIPS2024_b5b93943, muller2025sinq}.
Furthermore, these workloads tend to rely on numerous matrix multiplications, whose values are mostly concentrated around zero\cite{Gongyo_2024_ACCV}.
Consequently, the potential of finding better encodings for energy-efficient low-precision computations has become increasingly important.

% Unlike \gls{bo}/\gle{rl}/diffusion that optimize transformation sequences or parameterized \gls{rtl}, GENIAL directly searches operand encodings via surrogate inversion, with (i) a dedicated \gls{ssl} pretraining for encoding tensors, (ii) legality-aware \gls{ni} losses (attraction/repulsion), and (iii) a two-step EDA flow (lightweight evaluation\textrightarrow high-effort synthesis) that makes power-centric costly searches feasible.

Unlike prior approaches, such as Bayesian Optimization \cite{grosnit2022boils} or reinforcement learning \cite{hosny2020drills}, that optimize transformation sequences or parameterized \gls{rtl} templates, GENIAL directly targets the operand encoding itself. This is realized through a combination of (i) self-supervised pretraining to learn robust encoding representations, (ii) legality-aware network inversion losses that preserve bijectivity, and (iii) a hierarchical EDA-in-the-loop flow with options trading off execution time and fidelity.
% a hierarchical, three-tier EDA-in-the-loop flow that trades off execution time against power proxy fidelity
% old: a hierarchical, three-tier EDA-in-the-loop flow that balances ultra-fast screening with high-fidelity, power-centric evaluation.
% a two-stage EDA-in-the-loop flow that balances rapid exploration with accurate, power-centric evaluation.

%With the ever increasing amount of matrix-multiply intensive workloads being built and deployed every day
We propose GENIAL: a framework that enables fast and effective binary encoding discovery for optimizing circuit-level metrics such as power or area.
While we mainly focus on multiplier units, we also show that our framework generalizes to other circuit types.
This work brings forth the following contributions:

\begin{itemize}
    \item We introduce GENIAL, a machine learning-based \gls{dse} framework for the discovery of novel binary encodings that optimize arithmetic circuits for power efficiency.

    \item We perform comprehensive ablation studies to justify each component within the framework.
    
    \item We show that generation via Network Inversion drastically accelerates \gls{dse}, achieving superior sample efficiency compared to random or heuristic search.

    \item We deploy a high-effort \gls{lso} flow to maximize the usefulness of the surrogate model inherent to GENIAL.

    \item We conduct large-scale \gls{dse} experiments on 4-bit multipliers, where GENIAL reaches the Pareto front for \gls{swact} and area, and discovers a new 4-bit binary encoding whose energy efficiency closely matches that of the \gls{sme} format.

    \item We release the full GENIAL framework, including training scripts, surrogate models, and EDA integration utilities, to foster reproducibility and further research. The code is available at: \url{https://github.com/huawei-csl/GENIAL}.
    
\end{itemize}
 % - Framework for Generative DSE with NI
 % - Prove NI efficiency for DSE
 % - Reopen the encoding pandora's box
 % - Found new encoding
 % - Reach pareto front with GENIAL

\section{Related Works}
\label{sec:related_works}

\subsection{Multiplier Design and Data Formats}
Research on multiplier design has long centered on minimizing transistor count, reducing power consumption, and improving timing characteristics.
Techniques such as Booth encoding \cite{booth1951} and the Wallace tree \cite{wallace1964} were among the earliest to address efficient partial product handling, and remain influential in modern designs.
Over time, these ideas have evolved into higher performance variants, particularly through modifications to Booth's original method \cite{wen_chang_high_speed_booth}.
The choice of numerical representation also has a direct impact on \gls{swact}, which in turn affects dynamic power.
\gls{tc} is widely used for signed fixed-point arithmetic, but its carry chains can lead to increased bit transitions and higher dynamic power usage \cite{SwitchinginmultipliersMsc}. 
The \gls{sm} encoding, by comparison, mitigates this by toggling fewer bits when values are near zero \cite{waeijen2018}.
Beyond these standard formats, custom encodings tailored to arithmetic circuits have also been proposed to improve power efficiency in multipliers \cite{costa2001power}.

\subsection{Heuristic and Bayesian Optimization}
Early efforts in design space exploration and optimization relied on exhaustive enumeration and meta‐heuristics such as simulated annealing to explore small design spaces.
For instance, the binary switching algorithm, a method for optimizing mappings\cite{64657}, was applied to discover encodings that reduce bit error rates in communication systems \cite{schreckenbach2003optimized}.
More principled explorations come from Bayesian optimization (BO) with Gaussian‐process surrogates \cite{grosnit2022boils}; later, deep neural networks were introduced to scale BO to higher dimensions \cite{snoek2015scalable}.
The BANANAS framework further improved convergence by combining BO with a path‐based neural predictor \cite{white2021bananas}.

\subsection{Reinforcement Learning and VAE}
Deep reinforcement learning has been used to select effective sequences of logic‐synthesis transformations\cite{hosny2020drills}, and multi‐agent RL has been applied to multiplier designs\cite{feng2024gomarl, zuo2023rl}.
Variational autoencoders were also deployed for circuit interpolation\cite{song2024circuitvae}.

\subsection{Network Inversion}
In computer vision, deep generator models synthesize inputs that maximally activate neurons \cite{nguyen2016synthesizing}, and Plug \& Play Generative Networks perform iterative latent‐space optimization \cite{nguyen2017plug}.
Feature‐visualization techniques decode preferred stimuli \cite{olah2017feature}.
In protein design, inversion of AlphaFold’s structure predictor enables state-of-the-art protein sequence generation \cite{jumper2021highly, goverde2023novo}.
Collectively, these network inversion techniques demonstrate powerful generative capabilities and directly inspired GENIAL’s surrogate-based inversion approach for combinational logic.

\subsection{Other Hardware Optimization Methods}

Another notable solution is Diffusion\cite{NEURIPS2020_4c5bcfec}, a method originally used for image generation\cite{dhariwal2021diffusion, NEURIPS2020_4c5bcfec} and black‐box optimization\cite{krishnamoorthy2023diffusion}.
Diffusion for hardware \gls{dse} has already been adapted \cite{ren2025diffuse} and shows promising capabilities for \gls{dse}.
Finally, LLM- and agentic-based optimization has also recently attracted increasing interest.
FunSearch uses Transformer agents to guide code‐generation tasks\cite{romera-paredes_mathematical_2024}, and AlphaEvolve employs LLMs with evolutionary strategies to propose design transformations\cite{novikov2025alphaevolve}. 
Similarly, \cite{yao2024rtlrewriter} optimizes \gls{rtl} code.
Instead of text, Circuit Transformer models directly learn on a representation of the data structures being optimized \cite{li2024circuit, li2024logic}.
Although extremely promising, LLM‐based methods for low‐level hardware synthesis remain nascent and can be costly in their execution.

\section{Design Space Exploration Framework}
\label{sec:method}
\subsection{Overview}
\begin{figure}[ptb]
    \centering
    % \captionsetup[figure]{skip=-5pt} % Reduce the gap between the image and caption
    % \captionsetup[subfloat]{labelformat=empty} % Remove subfloat labels
    \includegraphics[width=1.0\linewidth]{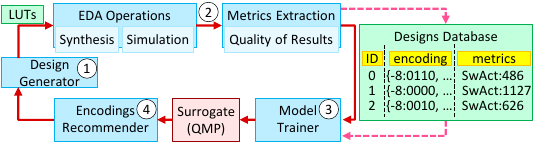}
    \caption{GENIAL framework overview: Generator \textrightarrow \  EDA Flows \textrightarrow \ Surrogate Model \textrightarrow \  Recommender}
    \label{fig:full_loop}
    % \vspace{-5pt}
\end{figure}

GENIAL is a framework aimed at exploring the space of operand encodings to minimize specific \gls{qor} metrics of a circuit, such as area or power.
As illustrated in Fig. \ref{fig:full_loop}, it consists of an iterative \gls{dse} process built with the following components:

\textbf{1. Design Generator:} converts the abstracted representation of a design to an \gls{rtl} description.

\textbf{2. EDA Task Launcher:} a server executing \gls{eda} steps such as synthesis, simulation, etc., to extract the \gls{qor} on all generated designs.

\textbf{3. \gls{qmp}:} a surrogate model trained to predict a target \gls{qor} metric for all designs out of their abstracted representations.

\textbf{4. Design Recommender:} a module that exploits the \gls{qmp} to recommend new designs to be processed next.

At each iteration, new designs (also called prototypes) are suggested, and further processed with \gls{eda} tools.
Relevant \gls{qor} metrics are extracted, and the \gls{qmp} is finetuned on the entire dataset acquired so far.
In this work, the term \textit{design} is used to refer to all logic circuits that have the same truth table, i.e., the same binary logic function.
The terms \textit{circuits} or \textit{graph} are used to refer to the different versions of the same design.

\subsection{Design Generator}
\label{sec:design_gen}
\subsubsection{Definition of Encoding}
\begin{figure}[htb]
    \centering
    % \captionsetup[figure]{skip=-5pt} % Reduce the gap between the image and caption
    % \captionsetup[subfloat]{labelformat=empty} % Remove subfloat labels
    \includegraphics[width=1.0\linewidth]{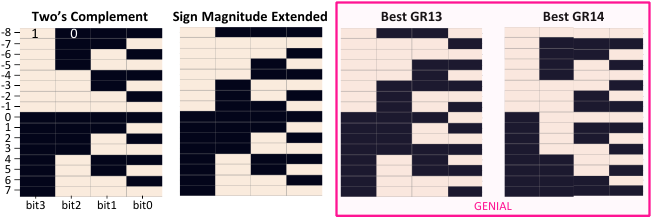}
    \caption{Representation of notable encodings as 2D tensors}
    \label{fig:design_representation}
    % \vspace{-5pt}
\end{figure}

An encoding is the mapping between a set of values and their associated binary representations.
E.g., in the \gls{tc} encoding, the binary representation of the value 3 is 0011 on 4 bits.
%An Encoding can be represented as a dictionary $v_i\mapsto c_i$ or as binary 2D tensors (Fig. \ref{fig:design_representation}).
A specific encoding can be specified as a dictionary $v_i\mapsto c_i$ from value to encoded representation, or as a codebook tensor (Fig.~\ref{fig:design_representation}).
%$C\in\{0,1\}^{2^n\times n}$ whose $i$-th row is the binary representation of $v_i$ (Fig.~\ref{fig:design_representation}).
There are $(2^4)!\approx 2.1 \cdot 10^{13}$ 4-bit binary encodings, which define the size of the search space for the 4-bit multiplier scenario.

\subsubsection{Abstracted Design Representation}
To interface with a machine learning pipeline, we must derive a representation of a design that can serve as input to a neural network.
Since our focus is on exploring different encodings that implement the same logic function (binary multiplication), we represent each design solely by its encoding, formatted as a tensor.
The columns correspond to the ordered individual bits of the binary representation, while the rows correspond to the ordered values being encoded. The \gls{tc} and \gls{sme} encoding representations are visible in Fig. \ref{fig:design_representation}.
% The values themselves are not directly.
% The actual values encoded do not matter but their ordering shall remain consistent.

\subsubsection{RTL Design Description}
% RA: Version prior to rebuttal
% Design generation is challenging as the \gls{rtl} must adhere to a set of rules to be synthesizable.
% To simplify this process, we represent each design as a \gls{lut} circuit, delegating all optimizations to the synthesis tool.
% Each design is generated by concatenating together the binary representations of the input operands into a selection vector, and assigning the output as the result of the corresponding operation. To ease integration with existing systems, the multiplier output is always in \gls{tc}, and we explore only custom input encodings.
% For instance, a 4-bit two-input and 8-bit single-output integer multiplier would be represented as follows in Verilog, assuming the \gls{tc} encoding in input:
% Design generation is challenging as the \gls{rtl} must adhere to a set of rules to be synthesizable.
% To simplify this process, we represent each design as a \gls{lut} circuit, delegating all optimizations to the synthesis tool.
% The LUT abstraction is essential in our context: while the inputs are expressed in arbitrary encodings obtained via automated exploration, in this work the outputs are always kept in \gls{tc} format to ensure seamless integration inside existing digital systems. The \gls{lut} representation thus provides a uniform and synthesizable way to guarantee correctness of the multiplication function for any input encoding, while the ouput encoding is always the two's complement.
Generating synthesizable \gls{rtl} for arbitrary operand encodings is non-trivial.
To simplify this process, each design is expressed as a \gls{lut} circuit, allowing downstream synthesis tools to handle structural optimization.
The \gls{lut} abstraction provides a technology-neutral, synthesizable representation that guarantees functional correctness for any input encoding, while outputs are consistently kept in \gls{tc} format to ensure compatibility with existing digital systems.
% Importantly, the LUT is not intended as the final hardware description; instead, it serves as a neutral intermediate form that enables encoding exploration.
% Optimized RTL implementations are then recovered through downstream synthesis flows, such as Yosys or FLOWY.
% For instance, a 4-bit two-input and 8-bit single-output integer multiplier would be represented as follows in Verilog, assuming the \gls{tc} encoding in input:
Importantly, the \gls{lut} is not the final hardware description but an intermediate form that decouples encoding legality from implementation details. Optimized \gls{rtl} implementations are subsequently recovered through reliable synthesis flows such as Yosys or FLOWY. For example, a 4-bit multiplier with two inputs and one 8-bit output can be represented as:

% \vspace{-2pt}
% \textcolor{red}{TODO: add the info that the output of the multiplier is always the TC!}

\begin{small}
\begin{verbatim}
    always @(sel) begin
        unique case(sel)
            ...
            8'b00100010 : out = 8'b00000100; // 2 x 2 = 4
            ...
        endcase
    end
\end{verbatim}
% \vspace{-5pt}
\end{small}

% LUT-based RTL is a technology-neutral intermediate representation to decouple encoding legality from implementation choices.
% The final netlists are obtained via reliable synthesis tools (Yosys/Mockturtle).
% LUTs are non-optimized, and can drastically bias power-related results (see Table \ref{tab:openr_flow_issue}).
% Hence, we thoroughly explore inter-flow metric correlations, and rely on the FLOWY synthesis results for the final results.
As illustrated, the two 4-bit operands are concatenated into an 8-bit input pattern. All $2^8$ input patterns are enumerated with their corresponding 8-bit outputs.

\subsection{EDA Task Launcher}
\label{sec:task_launcher}
% RA: Version prior to rebuttal
% The task launcher performs a sequence of EDA tasks on each generated design, including synthesis, post-synthesis simulation, power estimation, and finally \gls{qor} metric extraction.
% Leveraging different tools and flow scripts enables the extraction of various \gls{qor} metrics that capture key characteristics of a design.
% These metrics include \gls{ntrans}, post-synthesis simulation-based \acrlong{swact} or \gls{pow}, area footprint, number of \gls{mig} nodes (NMIG), and others.
% In the following, we present the different flows used in this work, which are also summarized Table \ref{tab:flows_summary}.
% RA: Version for rebuttal
The task launcher performs a sequence of EDA tasks on each generated design, including synthesis, power estimation, and finally \gls{qor} metric extraction. Crucially, different metrics have different trade-offs in usefulness vs. cost: for instance, \gls{swact} provides a close proxy to dynamic power, but requires full simulation, whereas simple complexity counts can be obtained nearly instantly. To balance accuracy and scalability, we implement three complementary flows and use them hierarchically: ESPRS for ultra-fast screening with \textit{\gls{cmplx}}, OPENR for mid-fidelity \textit{\gls{ntrans}}/\textit{\gls{pow}}, and FLOWY for final \textit{\gls{swact}}/\textit{\gls{ntrans}}.

% All this data is then stored in a database.
% The main flows used for this work are 1) ESPRS: the Espresso algorithm\cite{brayton_logic_1984}, 2) OPENR: OpenRoad flow scripts\cite{ajayi_openroad_2019}, which relies on Yosys\cite{wolf_yosys_nodate} for synthesis, and 3) FLOWY: a custom synthesis flow\cite{lee_late_2024, arnold_late_2025}.

% RA: Version prior to rebuttal
% % \subsubsection{Espresso Complexity Flow (ESPRS)}
% \subsubsection{Espresso-based Flow (ESPRS)} Espresso \cite{brayton_logic_1984} is a widely used, efficient logic minimization algorithm that reduces the size of a design's truth table.
% % Once a design is minimized, we can count the number of ANDs from the minterms and ORs from the maxterms, which yields a number that we define as \gls{cmplx}.
% %Once a design is minimized, we count the number of AND and OR gates, It can be used to estimate the \gls{cmplx} of a given design from its truth table.
% After minimization, we compute the complexity metric (CPX) by summing the counts of AND gates (in minterms) and OR gates (in maxterms).
% % We have observed that there is a very high correlation between the \gls{cmplx} and estimated \gls{ntrans} by Yosys from the second flow.

% RA: Version for rebuttal
\subsubsection{Espresso-based Flow (ESPRS, very fast/low fidelity)} Espresso \cite{brayton_logic_1984} is a well-established logic minimization algorithm that reduces the size of a design's truth table. After minimization, we compute the complexity metric (\gls{cmplx}) by summing the counts of AND gates (in minterms) and OR gates (in maxterms). Although coarse, \gls{cmplx} is strongly correlated with transistor count derived by Yosys in the OPENR flow (Pearson $\rho = 0.71$, see Fig. \ref{fig:correlation}), making it a lightweight proxy for key \gls{qor} metrics that can be used during design space exploration.

% RA: Version prior to rebuttal
% \subsubsection{OpenRoad Flow (OPENR)}
% The OpenRoad Flow Scripts\cite{ajayi_openroad_2019} are a fully open-source \gls{eda} framework that allow automated synthesis and place and route operations.
% Internally, it synthesizes the design using Yosys \cite{wolf_yosys_nodate} and performs place and route of the design using OpenROAD \cite{the_openroad_project_openroad_2025}.
% Combined with the OpenSTA \cite{the_openroad_project_opensta_2025} tool, this setup enables the extraction of delay and power information from the synthesized designs.
% % We have observed that the estimated number of transistors highly correlates with the estimated power when performing synthesis with Yosys, making it a good-enough proxy for validation of the method (namely running ablations, etc.).
% The primary metrics of interests in this flow are the \acrlong{ntrans} and \acrlong{pow}, with the former being faster to compute, as it does not require simulation.
% \gls{ntrans} is estimated using Yosys' internal transistor count model.
% \gls{pow} is obtained by mapping the design to the ASAP7 predictive Process Design Kit (PDK)\cite{CLARK2016105}, followed by simulation using the open source Verilator\cite{verilator} Verilog simulator (v0.53).

% RA: Version for rebuttal
\subsubsection{OpenROAD Flow (OPENR)}
The OpenROAD Flow Scripts \cite{ajayi_openroad_2019} provide a fully open-source pipeline for synthesis and place-and-route. Internally, Yosys \cite{wolf_yosys_nodate} performs the logic synthesis, while OpenROAD \cite{the_openroad_project_openroad_2025} and OpenSTA \cite{the_openroad_project_opensta_2025} handle placement, routing, and timing/power analysis. 
The primary metrics of interest in this flow are the \textit{\acrlong{ntrans}} (\gls{ntrans}) and \textit{average \acrlong{pow}} (\gls{pow}), with the former being faster to compute, as it does not require simulation.
\gls{ntrans} is estimated using Yosys' internal transistor count model.
To extract \gls{pow}, the design is mapped to the ASAP7 predictive Process Design Kit (PDK)\cite{CLARK2016105}, followed by simulation using the open-source Verilator\cite{verilator} simulator (v0.53). Finally, \gls{pow} is obtained using local \gls{swact} data combined with PDK information.
This flow strikes a balance between rapid evaluation and metric fidelity representative of hardware implementation.

% RA: Version prior to rebuttal
% \subsubsection{Custom Synthesis Flow (FLOWY)}
% Finally, we introduce a custom synthesis flow, highly inspired by the ones of \cite{lee_late_2024, arnold_late_2025}.
% It consists in an initial synthesis with Yosys, succeeded by a high effort \gls{lso} using the \gls{mt}\cite{riener_scalable_2019} tool by iterative random walk, as illustrated Fig. \ref{fig:flowy_pipeline}.
% For each generated \gls{lut}, the high effort \gls{lso} runs for 3000 steps on  12 parallel sequences of steps, called chains.
% At each step, a recipe is randomly selected from 30 available options and applied once if it is a decompression script, or three times if it is a compression script.
% At the end of the 3000 steps, the circuit with lowest number of \gls{mig} gates is selected from each chain.
% These 12 circuits are then simulated with Verilator, and the one exhibiting the lowest \gls{swact} is chosen as the final circuit for the \gls{lut}.

% RA: Version for rebuttal
\subsubsection{Custom Synthesis Flow (FLOWY)}
\begin{figure}
    \centering
    % \captionsetup[figure]{skip=-5pt} % Reduce the gap between the image and caption
    % \captionsetup[subfloat]{labelformat=empty} % Remove subfloat labels
    \includegraphics[width=1.00\linewidth]{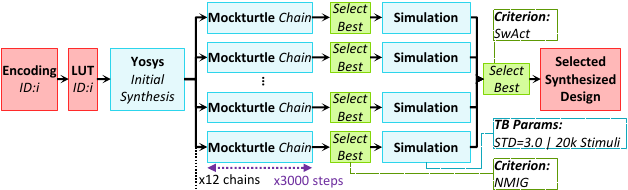}
    \caption{Custom high-effort logic synthesis optimization flow}
    \label{fig:flowy_pipeline}
    % \vspace{-5pt}
\end{figure}

We introduce a custom synthesis flow, highly inspired by the ones of \cite{lee_late_2024, arnold_late_2025}.
It consists in an initial synthesis with Yosys, succeeded by a high effort \gls{lso} using the \gls{mt}\cite{riener_scalable_2019} tool by iterative random walk, as illustrated in Fig. \ref{fig:flowy_pipeline}.
For each generated \gls{lut}, the high effort \gls{lso} runs for 3000 steps on  12 parallel sequences of steps, called chains.
At each step, a recipe is randomly selected from 30 available options and applied once if it is a decompression script, or three times if it is a compression script.
At the end of the 3000 steps, the circuit with the lowest number of \gls{mig} gates is selected from each chain.
These 12 circuits are subsequently simulated using Verilator, and the one with the lowest \gls{swact} is selected as the final implementation for the \gls{lut}.
% This process is significantly slower than ESPRS or OPENR, but it provides highly optimized transistor-level designs that approximate what a real EDA backend would produce.
This process is significantly slower than ESPRS or OPENR but yields highly optimized transistor-level designs that closely approximate the output of a real EDA backend.

\begin{table}[h]
    \centering
    \caption{Summary of Synthesis Flows and Associated Metrics}
    \begin{tabular}{|c|c|c|}
        Flow ID&Method&Metrics\\
        \hline
        ESPRS&Espresso&CPX\\
        OPENR&Open Road FlowScripts&NT, P\\
        FLOWY&Yosys + MT Random Walk&NT, NMIG, \gls{swact}\\
    \end{tabular}
    \label{tab:flows_summary}
\end{table}

\subsubsection{Flow Selection and Runtime Considerations}
\label{sec:flow_selection}

To balance accuracy and scalability, we employ a hierarchical approach using the three EDA flows introduced above:
\begin{itemize}[leftmargin=*]
    \item \textbf{ESPRS} (Fast Screening): Ultra-lightweight structural proxy, allowing rapid filtering of low-quality designs based on \gls{cmplx}.
    \item \textbf{OPENR} (Mid-Fidelity): Logic-synthesis pass evaluating normalized \gls{ntrans} and coarse \gls{pow}, used for downselection after ESPRS.
    \item \textbf{FLOWY} (High-Fidelity): Full LSO evaluation including \gls{swact}, applied only to finalist candidates.
\end{itemize}

This hierarchical strategy renders end-to-end search practical by evaluating only a small subset of promising designs with the more computationally expensive flow. Table~\ref{tab:flow_runtime} reports typical runtime per design evaluation on our hardware (8-core CPU@3.5\,GHz, no GPU acceleration for synthesis).

\begin{table}[h]
\centering
\caption{Typical runtime per design evaluation for the three EDA flows used in GENIAL.}
\label{tab:flow_runtime}
\begin{tabular}{l c c}
\toprule
\textbf{Flow} & \textbf{Runtime / sample} & \#CPUs\\
\midrule
ESPRS & $<$ 0.1\,s & 1\\
OPENR & $\approx$ 1\,min & 1\\
FLOWY & $\approx$ 15\,min & 12\\
\bottomrule
\end{tabular}
\end{table}

In practice, ESPRS evaluates the full set of candidates generated by Network Inversion, OPENR is applied to the top fraction ranked by the surrogate model, and only a small number of finalists proceed to FLOWY for accurate power-proxy estimation.

\subsubsection{Switching Activity Model}
For scalability, we introduce a lightweight power estimation protocol based on \gls{swact} extraction, described below.
% RTL simulation:

\begin{figure}[htbp]
\centering
\includegraphics[width=0.5\columnwidth]{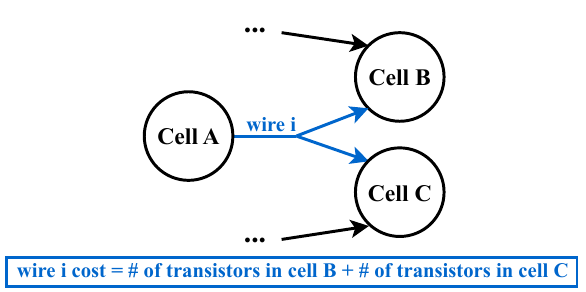}
\caption{\acrlong{swact} model for a cell with a fanout of two}
\label{fig:swact_model}
% \vspace{-5pt}
\end{figure}

\begin{enumerate}[label=\roman*., leftmargin=0.8cm]
    \item Stimulus generation:
sample an input, a 20k-long sequence of input value pairs from a normal distribution, and apply one pair of values to the multiplier’s input ports per clock cycle.
    \item State logging:
at each cycle, record the binary state of every input, output, and internal net.
    \item Transition counting:
count how many times each net state flips between consecutive cycles.
Multiply each flip count by the total transistor count of that net’s fan-out cells, thereby approximating the real energy cost of an individual wire's transitions.
This is illustrated in Fig. \ref{fig:swact_model}.
    \item Activity aggregation:
sum all weighted costs of all nets at each clock cycle, and average over the clock cycles to derive the circuit’s average \gls{swact}. %(reported here in arbitrary units). 
\end{enumerate}

% \textbf{(i)} Stimulus generation:
% sample an input, a 20k-long sequence of input value pairs from a normal distribution, and apply one pair of values to the multiplier’s input ports per clock cycle.

% \textbf{(ii)} State logging:
% at each cycle, record the binary state of every input, output, and internal net.

% \textbf{(iii)} Transition counting:
% count how many times each net state flips between consecutive cycles.
% Multiply each flip count by the total transistor count of that net’s fan-out cells, thereby approximating the real energy cost of an individual wire's transitions.
% This is illustrated Fig. \ref{fig:swact_model}.
% %Because output ports have no downstream cells, their fan-out weight is zero and they contribute nothing to \gls{swact}.

% \textbf{(iv)} Activity aggregation:
% sum all weighted costs of all nets at each clock cycle, and average over the clock cycles to derive the circuit’s average \gls{swact}. %(reported here in arbitrary units). 
% % \end{enumerate}

% This proxy allows fast evaluation of millions of designs, complementing the slower but more accurate FLOWS above.
We validate \gls{swact} as a proxy via strong correlations to Power and even Number of Transistors (Fig. \ref{fig:correlation}). Its use in FLOWY enables accurate selection of final designs without the need to fully simulate all candidates.

% Finally, for a two’s-complement multiplier with equal encoding, the overall activity is : $s_{tot}=2 \cdot s_{enc} + s_{mult} \label{eq:s_mm}$, where $s_{enc}$ and $s_{mult}$ are the encoder’s and multiplier’s \gls{swact}s, respectively.

\subsection{Quality Metric Predictor}
\label{sec:qmp}
\subsubsection{Quality of Results Prediction}
To recommend new designs, it is necessary to predict their \gls{qor}.
Accordingly, we train a deep learning network to perform a regression on the design scores.
The network takes as input the 2D tensor representation of a design's encoding and outputs a scalar value, which corresponds to the predicted score for that encoding.
Each known design is assigned a score, which is the standardized value of a target \gls{qor} metric.
The encoding-score pairs serve as supervised training samples.
% All known designs that have been processed by an \gls{eda} flow can therefore have an associated score label can be used as supervised training samples.
%(It takes in input an encoding representation, and delivers a "score" prediction, with the score being the standardized quality of result metric of a given design.)
%Only designs which have been processed by one of the \gls{eda} flows can be used as supervised training samples.
%(Only the designs which already went through one of the \gls{eda} flows are used for training.)

\subsubsection{Network Architecture}
\begin{figure}
    \centering
    % \captionsetup[figure]{skip=-5pt} % Reduce the gap between the image and caption
    % \captionsetup[subfloat]{labelformat=empty} % Remove subfloat labels
    \includegraphics[width=0.9\linewidth]{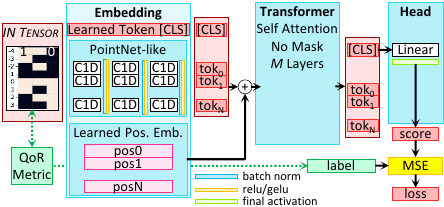}
    \caption{\acrlong{qmp} model architecture}
    \label{fig:predictor_network}
    % \vspace{-5pt}
\end{figure}

Fig. \ref{fig:predictor_network} depicts the architecture of the surrogate model, which consists in a PointNet-based\cite{Qi_2017_CVPR} embedding followed by a stack of Transformer encoder layers\cite{NIPS2017_3f5ee243}.

\textbf{Embedding:} The PointNet, a sequence of one-dimensional convolution layers (C1D), converts each individual binary representation of the input encoding into a fixed-size vector, also called a \textit{token}.
A learned vector of the same dimension, which serves as a global representation of the design for the Transformer model and commonly called the [CLS] token, is prepended to this sequence of vectors.
Finally, a learned sequence of absolute positional encodings is added element-wise to the token sequence.

\textbf{Transformer and Head:} The embedded token sequence is subsequently fed into a stack of Transformer encoder blocks.
Each block performs non-masked multi-head self-attention and outputs a new sequence of tokens.
At the output of the stack, only the vector corresponding to the [CLS] token of the final output sequence is passed through a series of linear layers, the model head, which outputs a single scalar value: the predicted score.

\subsubsection{Training Pipeline}
Training a network from random initialization is challenging, as it must learn both to efficiently represent each input vector in the embedding space and to predict \gls{qor} metrics from the resulting latent representation.
To accelerate convergence and improve final accuracy, we adopt a two-stage training pipeline, as well as a data augmentation scheme. 
% \textbf{(i)} \gls{ssl} pretraining to initialize meaningful embedding and attention weight, and 
% \textbf{(ii)} supervised finetuning on the metric regression task.

\textbf{Self-Supervised Pretraining:}
The \gls{ssl} pretraining phase consists of a simple permutation prediction task.
It involves predicting the positional indices of the permuted binary vectors.
The \gls{tc} encoding is used as the template encoding, with indices ordered from 0 to 15.
For each training sample, a random permutation of the 16 binary representations and associated indices is performed.
The permuted sample is passed through the Embedding and Transformer blocks.
A 2-layer perceptron (MLP) receives the [CLS] token output representation and predicts the positional indices corresponding to the permutation of the binary encodings.
The target is represented as a one-hot encoded sequence of length 16, and a cross entropy loss is applied.
Once the \gls{ssl} pretraining task is completed, the Embedding and Transformer weights are retained to initialize the model for finetuning, and the MLP head is discarded.

\textbf{Finetuning:}
In this phase, the model is trained on the real task of predicting \gls{qor} metrics.
The target score corresponds to the standardized \gls{qor} value of each known design, and the \gls{mse} serves as the loss function.
The network weights are optimized via backpropagation with the Adam optimizer \cite{kingma2017adammethodstochasticoptimization}, and the model learns to map an encoding input (represented as a 2D tensor) to its corresponding score.

\textbf{Data Augmentation:}
We exploit some notable properties of digital designs to derive a data augmentation scheme:
\begin{enumerate}[label=\roman*., leftmargin=0.8cm]
% \textbf{(i)} 
\item Swapping the positions of a logic unit’s input bits leaves its fundamental function unchanged.
Similarly, permuting the columns of the encoding tensor does not change the logic function of the associated design.
There are 24 possible permutations of 4 columns.

% \textbf{(ii)} 
\item Adding inverters in front of all input bits is essentially free in terms of added logic gates.
Furthermore, the \gls{swact} depends on signal toggles, and is independent of their polarity.
Hence, negating the encoding (by exchanging 0s and 1s) should likewise preserve the implementation cost.
\end{enumerate}

These two transformations preserve the legality of an encoding.
Thus each known design can be augmented with $2\times24=48$ different encodings resulting in the same \gls{qor}.

% Previous version felix:
%1) Swapping a logic unit’s inputs leaves its fundamental function unchanged. Colum permutation on the encoding tensor therefore yield equivalent designs in terms of logic complexity. 2) Adding input inverters is essentially free—both in transistor count and MIG-gates. Furthermore, \gls{swact} depends on signal toggles, not on their polarity. Hence, inverting columns likewise preserves implementation cost.
%Consequently, many encodings are considered equivalent under these transformations, and their QoR will be nearly identical. Therefore, we can augment our dataset by generating equivalent encodings and assigning each the same label as its original. It is also important to note that these transformations preserve legality

\subsection{Design Recommendation}
\label{sec:design_reco}
With the surrogate model finetuned, we leverage its predictions to guide the search toward promising designs and compare four different strategies for generating new candidates.
% The foundation of our framework relies on training a surrogate model able to predict a score that represents the quality of a design from its abstracted representation.

% Once the model weights have been trained as outlined above, there are multiple approaches to find new designs, which are discussed next.

\subsubsection{Naive Approach}
%From a trained score predictor model, a naive and trivial approach is to randomly generate a set of designs, predict their score with the network and select the best ones using a score threshold and apply an \gls{eda} flow on them to obtain their true QoR.
The \gls{na} is an intuitive baseline that involves:

\begin{enumerate}[label=\roman*., leftmargin=0.8cm]
\item randomly generating a large set of candidate encodings, 
\item predicting their score using the surrogate model, and 
\item selecting and processing only the top designs based on predicted scores.
\end{enumerate}

This process is repeated iteratively over several generations: at each round, newly generated designs are added to the existing dataset, and the surrogate model is finetuned with the updated data.
While this iterative refinement is expected to progressively guide the search toward higher quality designs (\gls{qor}), in practice, the performance of this approach remains significantly worse than the \gls{ni} approach described next.

\subsubsection{Network Inversion}
\begin{figure}
    \centering
    % \captionsetup[figure]{skip=-5pt} % Reduce the gap between the image and caption
    % \captionsetup[subfloat]{labelformat=empty} % Remove subfloat labels
    \includegraphics[width=0.7\linewidth]{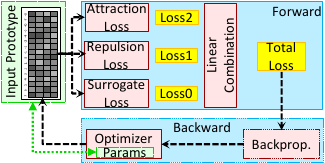}
    \caption{Network inversion setup}
    \label{fig:prototype_generation}
    % \vspace{-10pt}
\end{figure}

The \gls{ni} method employs a gradient-based optimization approach.
As illustrated in Fig. \ref{fig:prototype_generation}, it proceeds by:

\begin{enumerate}[label=\roman*., leftmargin=0.8cm]
\item initializing the input tensors with noise, 
\item performing a forward pass through the network,
\item backpropagating gradients to the inputs, and 
\item updating the inputs to minimize the loss.
\end{enumerate}

% Another method - called \gls{ni} - consists in 1) applying a forward pass on input tensors filled with noise, 2) backpropagating the gradients through the trained network up to the input, and 3) updating the input to minimize the loss of the network.
During optimization, the network weights remain frozen; only the input tensor, representing the design encoding, is updated using the Adam \cite{kingma2017adammethodstochasticoptimization} optimizer.
The primary component of the loss is the surrogate model’s output itself, since the objective is to identify input encodings that yield low \gls{qor} values.
The optimizer iteratively adjusts the input representation to reduce the loss, decreasing the surrogate prediction value.
To ensure that distinct designs are generated, each input tensor is initialized with a constant value of 0.5 and perturbed with small Gaussian noise.
This initialization strategy enables the generation of multiple distinct encodings while using the same fixed model parameters.
% Starting representations are tensors filled with 0.5 to which are added small random gaussian noise to generate different encodings out of the same model weights.
%One constraint that must be guaranteed is that the input encoding is valid.
%An input encoding is legal only if all the values are represented by unique bit representations of the right length).

An input encoding tensor is considered \textit{legal} if it contains every possible binary representation exactly once.
To increase the likelihood of generating such legal encodings, we introduce two regularization terms:

\begin{enumerate}[label=\roman*., leftmargin=0.8cm]
\item  An \textit{attraction loss}, which pulls each input vector toward all possible binary target vectors, encouraging convergence to binary values (0s and 1s).
\item A \textit{repulsion loss}, which pushes input vectors away from one another, discouraging vectors from converging to the same target.
\end{enumerate}
% that attracts all input vectors to all possible target vectors.
% This increase the probability to converge towards binary vectors (zeros and ones only).
% 2) A "repulsion" loss that repels all input vectors from each other.
% This reduces the chance of vectors converging to the same target vector.

These regularization terms are weighted by a scalar coefficient that balances the trade-off between minimizing the \gls{qor} loss and enforcing the attraction-repulsion constraints.
After 1500 optimization epochs, most of the vectors of an input converge to valid binary values, though some remain in intermediate states.
A post-processing step is thus applied to legalize the encoding.
Specifically, for each non-binary vector, we compute its distance to the remaining unassigned binary target vectors.
The vector-target pair with the smallest distance is then matched, and both are removed from their respective candidate pools.
This process is repeated iteratively until all vectors are assigned, yielding a fully legal encoding.
% For this, we measure the distance of the floating vectors to the remaining (non-assigned) target vectors.
% The floating with the minimum distance to any remaining target vector has the highest priority and gets assigned the according target vector.
% Both are removed from the list of non-assigned vectors and this is pursued until all list are cleared.

\section{Experiments and Results}
\subsection{Experimental Setup}
\label{sec:results_cmplx}

% We do a first random generation of N designs.
% Then we process all designs to get an initial dataset.

% We pre-train a model with about 200 epochs.

% We finetune it for about 200 to 400 epochs and we keep the model with the lowest validation loss.

% We then generate prototypes out of noise around grey 2D tensor, for about 1500 epochs.

% Those prototypes are then converted to RTL.
% And we start again.

The experiment involves the following steps.

\textbf{Model Pretraining:} The model is pretrained using \gls{ssl} for 200 epochs.

\textbf{Initial Generation:} A set of random valid designs is generated and processed to extract the QoR, forming the initial dataset.

\textbf{Model Finetuning:} The model is trained on the full available dataset for 200-400 epochs.
The checkpoint with the lowest validation loss is selected for the recommendation.

\textbf{Prototype Generation:} Recommended candidate encodings are generated via \gls{ni} for 1500 epochs. These 1500 \gls{ni} epochs yield \>99\% vectors binary-converged before legalization; beyond that we saw diminishing returns.

\textbf{QoR Extraction:} The designs corresponding to the selected encodings are generated as LUTs.
They are synthesized with one of the available flows, and their \gls{qor} are extracted.

\textbf{Iteration:} The newly evaluated designs are added to the dataset, and the process is repeated for another round.
At each generation round, the dataset used for finetuning the network is expanded with additional designs, leading to two key effects: it improves the maximum attainable accuracy of the surrogate model, and it enriches the dataset with more diverse design knowledge.
Consequently, surrogate-model-based recommendations generally improve over successive generation rounds.

\textbf{Design Methodology Reference:} 
For clarity, we briefly restate the role of each component in GENIAL:

\begin{enumerate}[label=\roman*.]
\item the \textit{Design Generator} converts candidate encodings into synthesizable RTL (Section \ref{sec:design_gen});
\item the \textit{EDA Task Launcher} evaluates each RTL design using one of the three flows described in Section \ref{sec:task_launcher};
\item the \textit{\acrlong{qmp}} predicts the \acrshort{qor} of new designs from their encoded representation (Section \ref{sec:qmp});
\item the \textit{Design Recommender} uses \acrlong{ni} to propose new encodings based on surrogate gradients (Section \ref{sec:design_reco}).
\end{enumerate}

\begin{figure}
    \centering
    \includegraphics[width=1.00\linewidth]{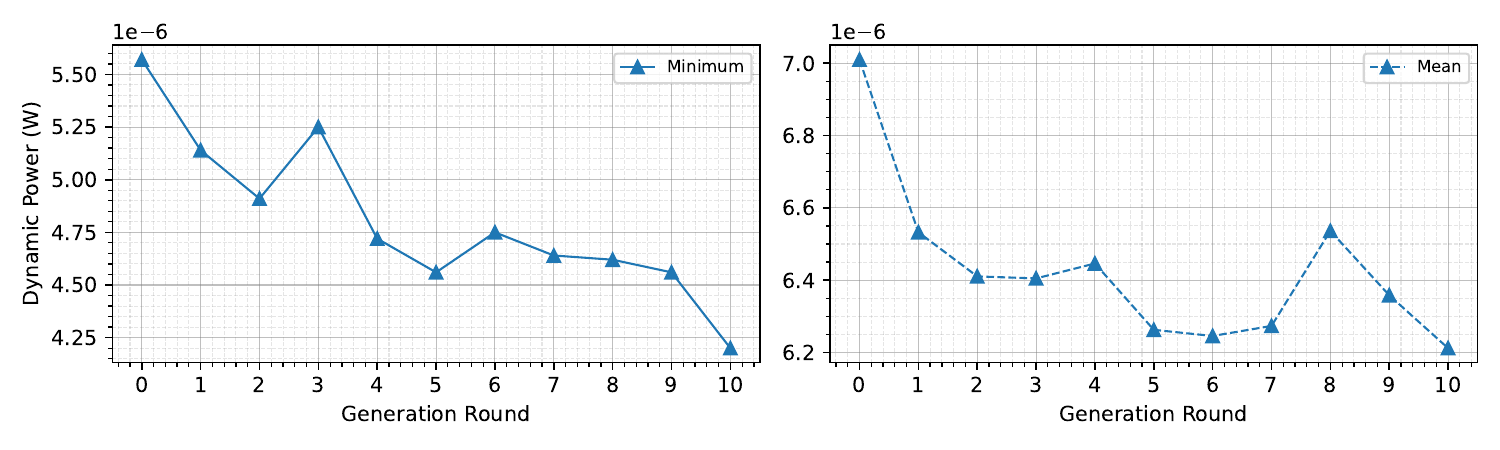}
    \caption{GENIAL results: OPENR with target Power (P)}
    \label{fig:power_loop_results}
    % \vspace{-10pt}
\end{figure}

\subsection{OPENR Optimization for Power}
\label{sec:results_ntrans}
In this experiment, we run GENIAL using the OPENR flow, targeting Power (P) as the \gls{qor} metric to optimize.
Results are obtained by synthesizing and simulating each design with OpenROAD Flow Scripts in the synthetic ASAP7 technology \cite{CLARK2016105}.
Each design is compiled and simulated with Verilator over 20k randomly generated stimuli (i.e. input operand pair switches), with a normal distribution of mean 0 and standard deviation ($\sigma=3$).
The values are clipped to the range (-8, 7) and rounded to the nearest integer.
Simulation waveforms are subsequently used to perform power extraction.

Fig. \ref{fig:power_loop_results} illustrates the evolution of the minimum and mean power (P) across generation rounds, showing that GENIAL consistently reduces power consumption over iterations.
However, the \gls{lut}-based \gls{sme} and \gls{tc} encodings were not optimized as effectively as their hand-crafted counterparts implemented with Verilog arithmetic operators, as shown in Table \ref{tab:openr_flow_issue}.
The LUT \gls{sme} design exhibits higher power than the LUT \gls{tc}, which should not be the case, as highlighted by power extraction results realized on handmade designs (see Table \ref{tab:openr_flow_issue}). 
Since GENIAL's ability to discover meaningful designs strongly depends on the accuracy of the \gls{qor} metrics used to train the \gls{qmp}, a higher-effort \gls{lso} in the loop is necessary.

\begin{table}[htb]
    \centering
    \caption{OPENR Synthesis and Power Extraction Results}
    \label{tab:openr_flow_issue}
    
    \begin{tabular}{r|rr|rr}
    \toprule
        \multirow{2}{*}{Metric}&\multicolumn{2}{c|}{Handmade}&\multicolumn{2}{c}{LUT}\\
                &TC&SME&TC&SME\\
        \midrule
        Area (\textmu m\textsuperscript{2})&6.64&8.1&17.8&25.7\\
        Power (\textmu W)&3.25&3.14&4.52&5.34\\
        Power vs. TC	&1.00	&0.97&	1.00&	1.18\\
        Max Delay (ps)&1.293&1.351&1.182&1.167\\
        \bottomrule
    \end{tabular}
    % \vspace{-8pt}  
\end{table}

\textbf{Note on LUT vs. handmade RTL.}
As seen in Table~\ref{tab:openr_flow_issue} (OPENR), LUT-\gls{sme} exhibits higher \gls{pow} than LUT-\gls{tc}, whereas handmade operator-based RTL shows the opposite trend.
We therefore treat LUT-\gls{rtl} power as a screening signal only and rely on FLOWY (high-effort \gls{lso}) and \gls{swact} for final conclusions in Sec.~\ref{sec:results_swact}.

\subsection{FLOWY Optimization for Switching Activity}
\label{sec:results_swact}
\begin{figure}
    \centering
    \includegraphics[trim=10 0 40 0,clip,width=1.00\linewidth]{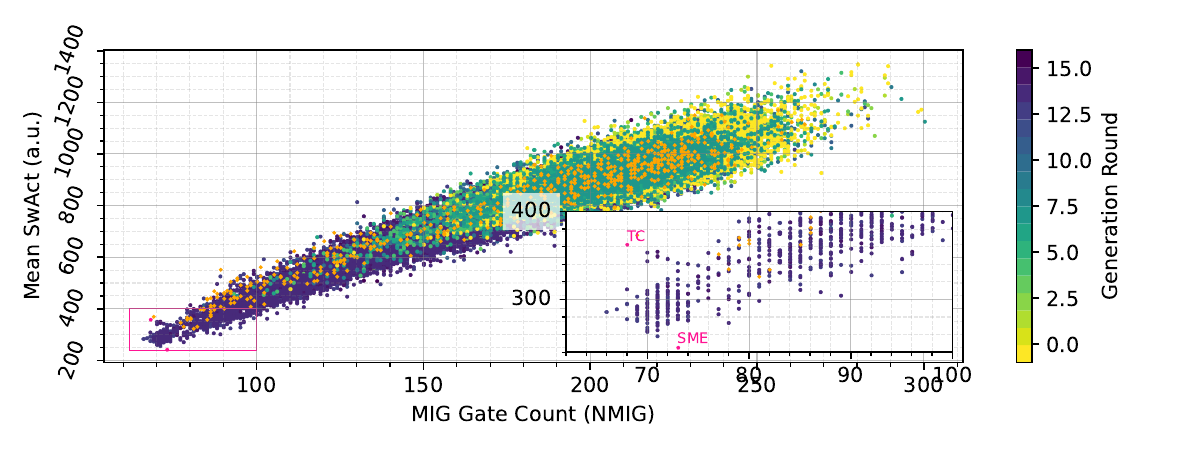}
    \caption{GENIAL results: FLOWY with target SwAct}
    \label{fig:flowy_loop_results}
    % \vspace{-5pt}
\end{figure}

We conduct a larger experiment with FLOWY in the loop, and the results are shown in Fig. \ref{fig:flowy_loop_results}.
The initial dataset consists of 10k randomly generated designs, augmented with 577 additional designs uniformly sampled from a separate dataset previously generated by running GENIAL with OPENR targeting \gls{ntrans} as the \gls{qor} metric.
This augmentation increases dataset diversity and maximizes the likelihood of successful optimization.
To avoid biasing the search, the initial dataset excludes both the \gls{tc} or \gls{sme} encodings.
We apply all proposed quality-enhancement techniques, including \gls{ssl} pretraining, data augmentation, and over-generation during \gls{ni}, followed by a selection of the top 25\% predicted encodings.
For the first 10 generation rounds, 2k designs are recommended per iteration.
Later, as training time became the limiting factor relative to design synthesis, we increased the number of designs per round to 4.5k.
The final simulation for each best design in every chain was performed with 20k stimuli, with operand values randomly sampled from a normal distribution with $\sigma=3.0$, as above.

\begin{figure}
\vspace{5pt}
\noindent
\begin{minipage}{0.5\linewidth}
    \includegraphics[width=\linewidth]{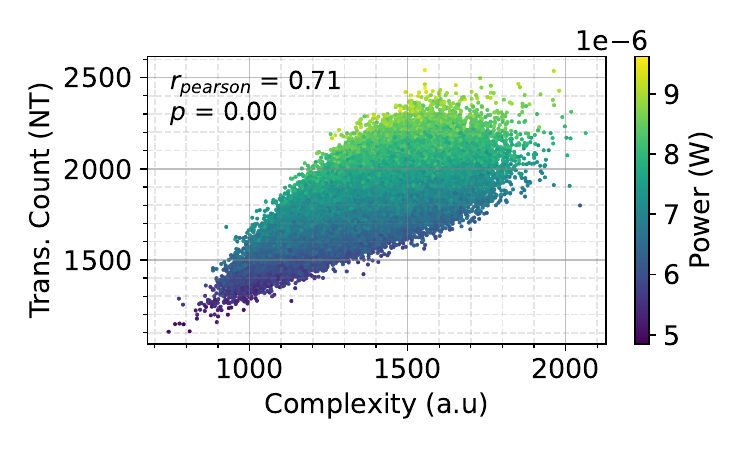}
\end{minipage}%
\begin{minipage}{0.5\linewidth}
    \includegraphics[width=\linewidth]{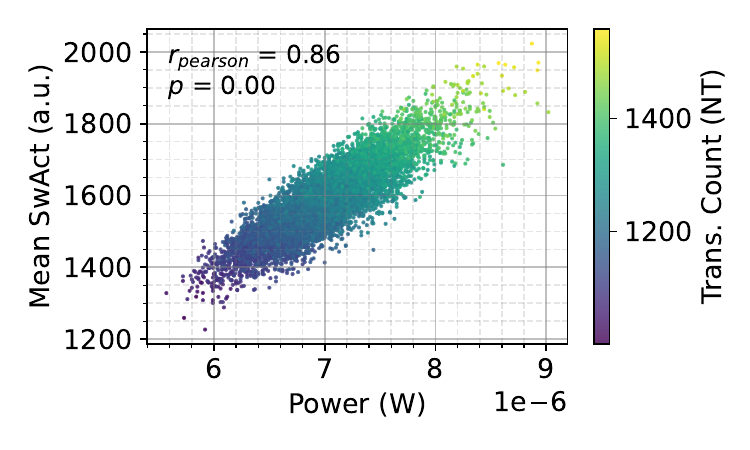}
\end{minipage}
\caption{Correlation between various target metrics}
\label{fig:correlation}
%\vspace{-10pt}
\end{figure}

After generation round 13 (GR13), we observe convergence towards an encoding that exhibits significant symmetry around zero, which is different from the \gls{sme}.
A similar encoding was found at generation round 14 (GR14).
When referring to GR13/GR14 encodings in Fig.~\ref{fig:flowy_loop_results}, we visualize them in Fig.~\ref{fig:design_representation}; correlations among proxies are in Fig.~\ref{fig:correlation}.

% All data points are illustrated Fig. \ref{fig:flowy_loop_results}, and the two encodings mentioned here are depicted Fig. \ref{fig:design_representation}.
% The encodings discovered at GR13/GR14 (cf. Fig. \ref{fig:design_representation}), exhibit symmetry around zero with reduced Hamming transitions near small magnitudes, differing from \gls{sme} while achieving competitive \gls{swact} (Fig.~\ref{fig:flowy_loop_results}).

We find several solutions at the Pareto-front of \gls{swact} vs \textit{NMIG}.
Particularly, we find solutions expected to be both smaller and to use less power than \gls{tc}.
Some encodings also use 12\% less area than SME for only 13\% more power.

We then perform a higher effort \gls{lso} (with 15k steps per chain and 120 chains per design) on selected encodings and present the results in Table \ref{tab:high_effort_synth} (a).
They show that the obtained encoding performs nearly as well as the \gls{sme} encoding and consumes 18\% less power than the \gls{tc}.
% obtained is nearly on par with SME and better than the two's complement one.

\begin{table}[h]
    \centering
    \caption{FLOWY Higher Effort Synthesis Results ($\sigma=3$)}
    \begin{tabular}{rr|rr|rrrr}
    % \begin{tabular}{|c|c||c|c||c|c|c|c|}
    \toprule
    \multicolumn{1}{c}{}&\multirow{2}{*}{Metric}&\multicolumn{2}{c|}{Handmade}&\multicolumn{4}{c}{LUT}\\
        
        \multicolumn{1}{c}{}&&TC&SME&TC&SME&GR13&GR14\\
        \midrule
    \multirow{4}{*}{\rotatebox{90}{Mult. (a)}}&NMIG&64&69&64&71&70&70\\
    &SwAct&286&231.3&321.0&235.0&263.4&272.2\\
    &vs. TC&1.00&\textbf{0.73}&1.00&\textbf{0.73}&0.82&0.85\\
    &Depth&9&12&11&9&9&9\\
        \midrule
                    
    \multirow{3}{*}{\rotatebox{90}{Enc. (b)}}&NMIG&-&-&0&8&8&8\\
            &SwAct&-&-&0.0&34.5&38.1&39.8\\
            &Depth&-&-&0&3&3&3\\
    \bottomrule
    \end{tabular}
    \label{tab:high_effort_synth}
    % \vspace{-2pt}
\end{table}

Note that the current code implementation (available at \href{https://github.com/huawei-csl/GENIAL}{https://github.com/huawei-csl/GENIAL}) is built around Docker containers and supports multi-node, multi-CPU parallelism. This allows the search to be conducted at scale within a reasonable time frame, even when using the high-effort FLOWY synthesis (approximately one week for the full run shown in Fig. \ref{fig:flowy_loop_results}).

% We compare out methods to other baselines. In particular, we set the number of allowed encoding sample evaluations (running the synthesis) and for each method log the minimum obtained value. Multiple runs are performed to get the expected minimum for each method. The results are shown in Fig. \ref{fig:method_comparision}. For both metrics evaluated - Espresso nodes and Yosys transistor count - our method greatly outperforms other approaches except for small sample sizes, which can be attributed to the minimum amount of samples required to train the prediction model. This results highlights the sample efficiency of our method. The included methods are A) Random Search: legal encodings are sampled randomly, B) Binary Switching Search: We start with a random legal encoding and evaluate the cost of all its swap neighbors, that is an encoding which can be obtained by only swapping two rows in the input representation (for 4 bit there are 15+14+..+1=120 swap neighbors). We continue with the swap neighbor with the lowest cost. Additionally, is a 25\% chance that we pick a random neighbor instead. After 5000 cost evaluations the search starts again from a random encoding. C) Naive Prediction: the approach outlined above. D) Ours: our method including, model is trained after each data point in the figure.

% shorter version

\section{Validation of the Method and Discussion}
\subsection{Metric Correlations}
\label{sec:metric_correlation}
GENIAL is rigorously validated through a series of experiments.
Since FLOWY is relatively slow (processing a single encoding takes about 15-20 min on 18 CPUs), we evaluate our methods using the faster ESPRS or OPENR flows.
We observe a strong correlation between most metrics across the different flows, as shown in Fig. \ref{fig:correlation}, confirming that ESPRS and OPENR flows are reliable proxies for tuning our method.

\begin{figure}
    \centering
    \includegraphics[width=1.00\linewidth]{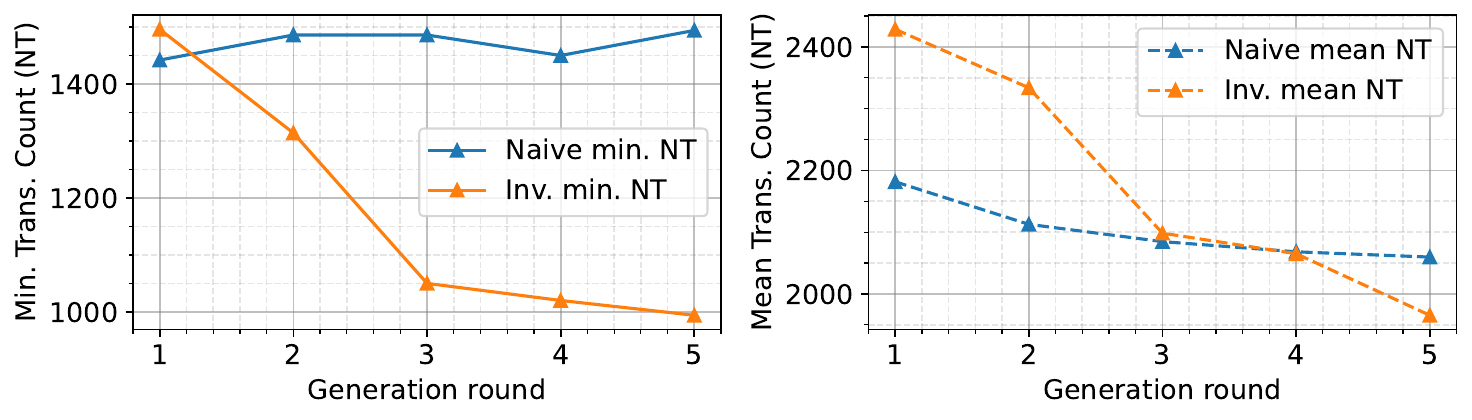}
    \caption{Comparison of Naive approach and Surrogate Model Inversion}
    \label{fig:naive_vs_inversion}
    %\vspace{-5pt}
\end{figure}

\subsection{Comparison with Other Methods}
\subsubsection{Naive versus Model Inversion}
To compare \gls{ni} against the \gls{na}, we evaluate both methods under equivalent compute duration, over three trials.
Since \gls{ni} requires several backward passes to converge on meaningful encoding prototypes, generating 10k samples takes approximately 25 minutes.
Therefore, for each generation round, the \gls{na} method was run for 25 minutes, resulting in about 10M samples.
% which results in the generation and \gls{qmp} forward pass of about 10M samples.
From these, only the top 10k designs based on predicted scores by the \gls{qmp} were retained and processed.
% Out of these, we keep only the 10k with the best score, and process them.
Results of this experiment, using the OPENR flow with \gls{ntrans} as the target \gls{qor} metric, are shown in Fig. \ref{fig:naive_vs_inversion}.
% Results for this experiment with OPENR and target transistor count (NT) are illustrated Fig. \ref{fig:naive_vs_inversion}
After just two generation rounds (30k processed samples), \gls{ni} already discovers encodings with lower minimum \gls{ntrans} than \gls{na}.
% It shows that in only two generation rounds (30k processed samples), \gls{ni} already finds encodings with NT count than \gls{na}.
After five rounds (60k processed samples), the \gls{ni}-recommended population has a lower mean \gls{ntrans} than the naive approach, demonstrating the superior effectiveness of \gls{ni}.

\begin{figure}
    \centering
    \includegraphics[width=1.00\linewidth]{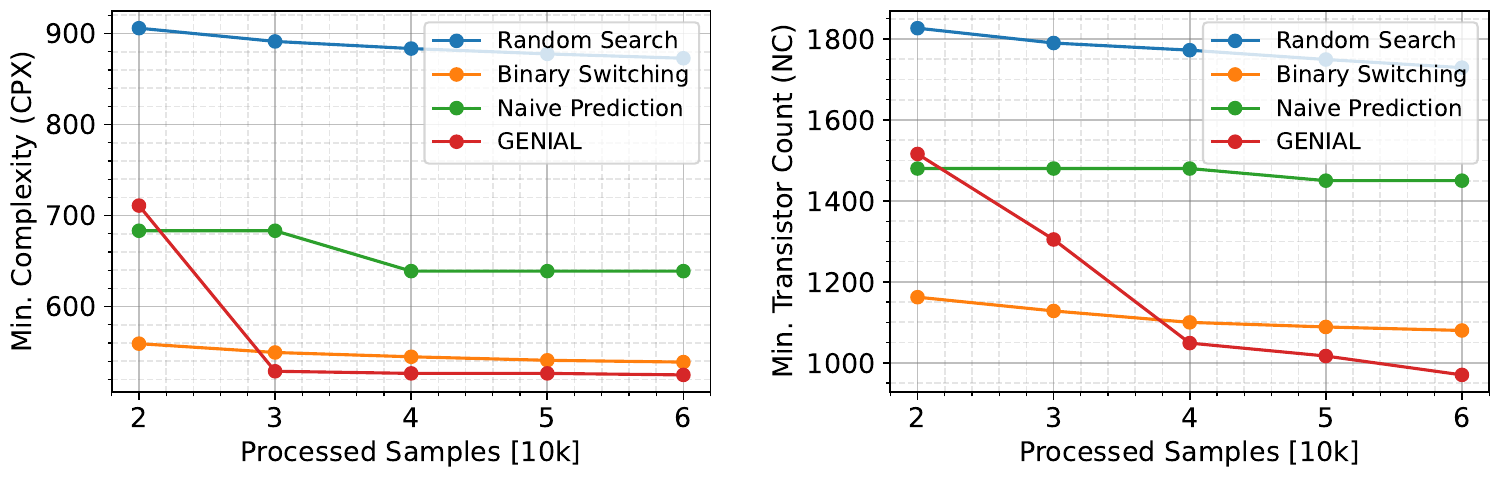}
    \caption{Comparison between methods presented in \cref{sec:sampleeff}.}
    \label{fig:method_comparison}
    % \vspace{-10pt}
\end{figure}

\subsubsection{Sample Efficiency}\label{sec:sampleeff}

% We compare the proposed Network Inversion approach with three other technics, and show that GENIAL achieves faster minimum than ...
We compare our method against three others by limiting the number of processed samples across three trials.
Fig.~\ref{fig:method_comparison} shows the expected minimum value achieved as a function of the sample budget for two metrics: \textit{CPX} and \textit{NT}.
%\newpage
The methods compared are:

\begin{enumerate}[label=\roman*., leftmargin=0.8cm]
\item Random Search: uniformly sample legal encodings.
\item Binary Switching Search\cite{64657}: starting from a random legal encoding, evaluate all \textit{swap neighbors} (encodings obtained by swapping two rows in the input representation), move to the lowest‐cost neighbor (with a 25\% random move chance), and restart from a new random encoding every 5000 samples processed.
\item Naive Approach with Surrogate Models: as described in Section~\ref{sec:design_reco}.
\item GENIAL: using 10k initial samples and 10k samples per generation.
\end{enumerate}

Except at very low budgets (where the network has insufficient samples to effectively train its parameters), GENIAL consistently finds the lowest-cost encodings, demonstrating superior sample efficiency.
At an equal number of processed samples, \gls{ni} consistently attains lower minima and means than \gls{na} because it exploits surrogate gradients to target promising encoding manifolds directly, rather than relying on random sampling plus predictions (Fig.~\ref{fig:method_comparison}). As the dataset grows, the \gls{qmp}'s accuracy improves and \gls{ni}'s sample efficiency further increases.%

\subsection{QMP Training Ablations}
\label{sec:ablations}
\subsubsection{SSL Pretraining}
\begin{figure}
\noindent
\begin{minipage}{\linewidth}
    % Top row: two side by side, inside this minipage
    \begin{minipage}{0.495\linewidth}
        \includegraphics[width=\linewidth]{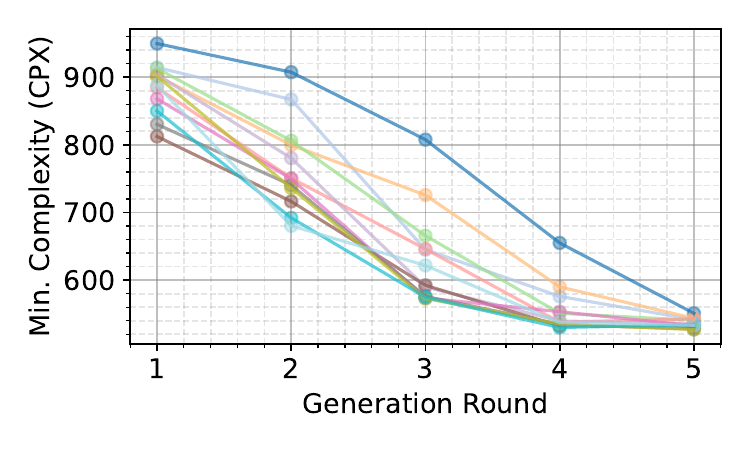}
    \end{minipage}%
    \begin{minipage}{0.495\linewidth}
        \includegraphics[width=\linewidth]{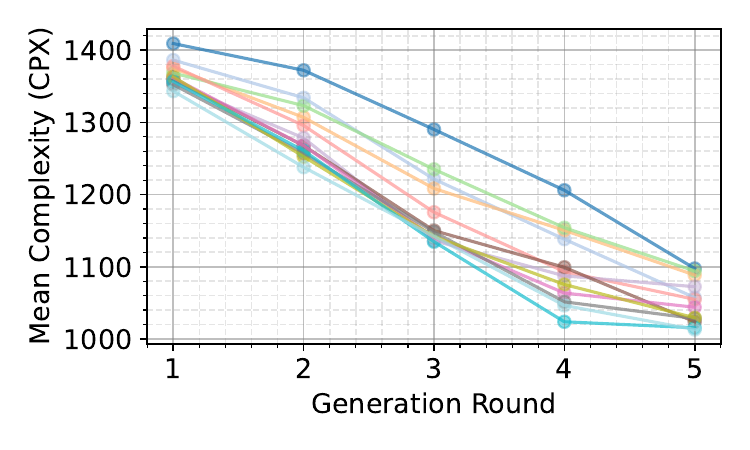}
    \end{minipage}
    
    % Bottom row: one spanning the full width
    \begin{minipage}{\linewidth}
        \includegraphics[width=\linewidth]{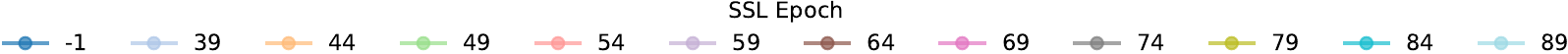}
    \end{minipage}
\end{minipage}
\caption{Ablation study results: pretraining efforts}
\label{fig:ssl_ablation}
%\vspace{-5pt}
\end{figure}

Optimizing with ESPRS, we assess the impact of \gls{ssl} pretraining on recommendation quality by performing one generation round with different model checkpoints, each picked at a specific epoch during \gls{ssl}.
Each data point in Fig. \ref{fig:ssl_ablation} is the average of three independent runs.
While higher epochs generally lead to better recommendations, no clear optimal epoch emerges.
% It shows that a higher \gls{ssl} epoch generally results in better recommendations.
% There is however no clear optimal epoch value.

% \begin{figure}
%     \centering
%     \includegraphics[width=1.00\linewidth]{aspdac26_paper/figures/fsm_transistors_loop_paper.pdf}
%     \caption{Ablation study results: pretraining \textcolor{red}{TODO:change graph! (currently FSM)}}
%     \label{fig:ssl_ablation}
%     \vspace{-5pt}
% \end{figure}

% \begin{figure}
% \noindent
% \begin{minipage}{0.5\linewidth}
%     \includegraphics[width=\linewidth]{aspdac26_paper/figures/ssl_checkpoint_cmplx/plot_min_complexity.pdf}
% \end{minipage}%
% \begin{minipage}{0.5\linewidth}
%     \includegraphics[width=\linewidth]{aspdac26_paper/figures/ssl_checkpoint_cmplx/plot_mean_complexity.pdf}
% \end{minipage}
% \vspace{-5pt} % (optional: tweak for vertical gap)
% \begin{minipage}{\linewidth}
%     \includegraphics[width=\linewidth]{aspdac26_paper/figures/ssl_checkpoint_cmplx/legend_ssl_epoch_new_cropped.pdf}
% \end{minipage}
% \caption{Ablation study results: pretraining efforts}
% \label{fig:ssl_ablation}
% \vspace{-5pt}
% \end{figure}

\subsubsection{Data Augmentation}
\begin{figure}
    \centering
    \includegraphics[width=1.00\linewidth]{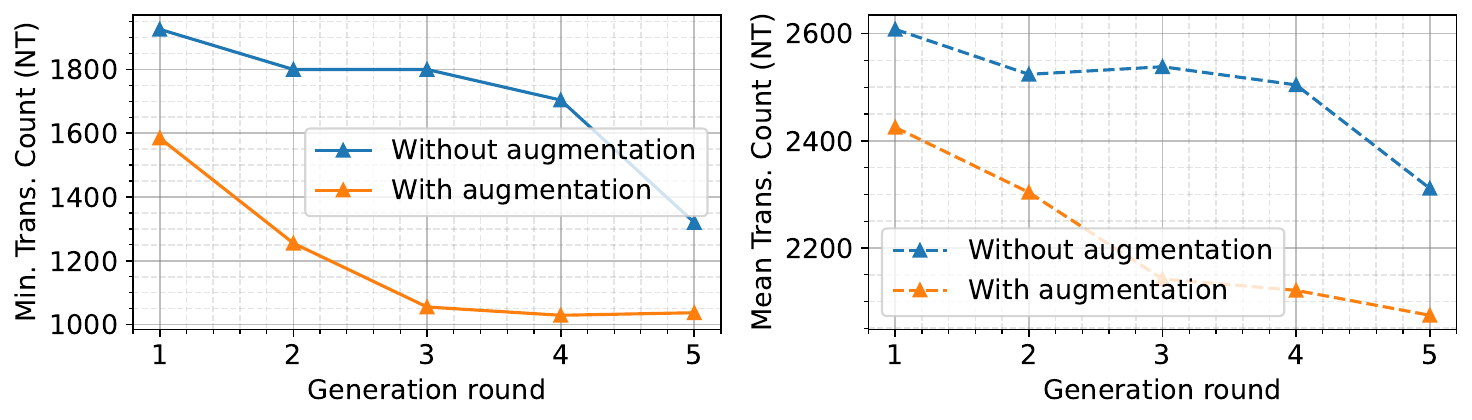}
    \caption{Ablation study results: data-augmentation}
    \label{fig:augmentation_comparison}
    % \vspace{-5pt}
\end{figure}

Using OPENR for optimizing \gls{ntrans}, we assess the impact of \gls{da} during finetuning by running two experiments starting from the same dataset and pretrained models.
As shown in Fig. \ref{fig:augmentation_comparison}, \gls{da} significantly improves recommendation quality, with both the minimum and mean \gls{ntrans} consistently lower across all generation rounds compared to the baseline without \gls{da}.

% \subsubsection{Variational Auto Encoders}
% \textcolor{olive}{\textbf{Should we keep that?}:
% We evaluate the approach proposed in \cite{song2024circuitvae} by adding the VAE constraint on the learned token $tok_L$ latent space right before the model's head (see Fig. \ref{fig:predictor_network}) during \textcolor{red}{@Ryan: both? pretraining and finetuning.}.
% Optimizing with OPENR and target NT, we do not observe any performance improvement when the VAE constraint is applied.}
\vspace{-2pt}
\subsection{Other Considerations}
\label{sec:other_designs}
\subsubsection{Optimizing Finite State Machines}
\begin{figure}
    \centering
    \includegraphics[width=1.00\linewidth]{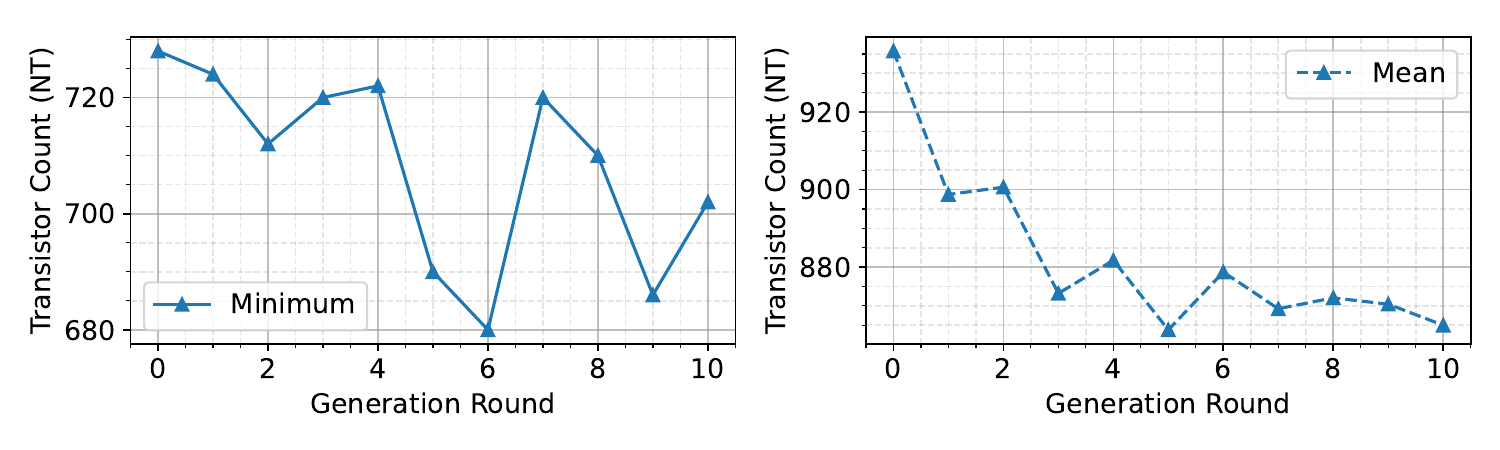}
    \caption{Optimization results for a finite state machine}
    \label{fig:fsm_results}
    \vspace{-10pt}
\end{figure}

% In order to validate the generality of our method we apply it to optimize state machines which apart from arithmetic units is another type of frequently occurring logic blocks. In particular we implement a Moore type state machine with 16 states (4 bits to encode a state), 3 input bits and 2 output bits. The state machine's transitions (15 clauses) and output decoder is automatically randomly generated with a script. We employ our method (without data augmentation) with 30k initial dataset and 5k prototypes per generation. The result is shown in Fig. \ref{fig:fsm_results}.

% shorter version
To evaluate the generalizability of our method, we applied it to Moore-type \gls{fsm}, a common logic block beyond arithmetic units.
We randomly generated a 16-state \gls{fsm} (using 4-bit state encodings) with 3 inputs, 2 outputs, and 15 transition clauses.
Starting from a random 30k-sample initial dataset and producing 5k encodings per GR with GENIAL, we observe that the minimum and mean \gls{ntrans} are significantly lower than the initial dataset (GR0) (see Fig. \ref{fig:fsm_results}). 

Note that GENIAL is built for encoding discovery, and therefore cannot be applied to any random logic designs.
GENIAL specifically targets problems with optimizable encodings.

% Running our method (without \gls{da}) to optimize the state encoding with a 30k-sample initial dataset and 5k prototypes per generation yields the results shown in Fig. \ref{fig:fsm_results}.
% The prototype generations (rounds 1-10) exhibit significantly lower minimum and mean values than the initial random dataset (round 0).

\subsubsection{System-level considerations}
While optimizing the operand encoding of a multiplier can reduce its power consumption, the impact must be assessed at the system level.
Specifically, inputs to the multiplier, typically represented in \gls{tc}, would need to be converted to a custom encoding used by the multiplier.
This conversion is performed by an encoder (Enc.).
Their size and cost are visible in Table \ref{tab:high_effort_synth} (b).
However, if the same operand can be converted once and reused multiple times, as is common in neural network accelerators, the cost of this conversion is amortized by the reuse factor.

Finally, we validated the study by implementing a DaVinci-like core \cite{cube}, which realized a General Matrix Multiplication (GEMM) operation (\textit{cube}) containing $16\times32\times16$ multipliers.
We used commercial tools for synthesis, back-annotation for glitch-aware simulation, and power extraction.
By swapping out the multipliers and placing the encoders right before the matrix buffers (L0), we observed major gains:

\begin{enumerate}[label=\roman*., leftmargin=0.8cm]
    \item SME multipliers enable up to 16\% power savings against their two's complement counterpart on a 2D convolution operation, where both operand matrices are filled with values randomly sampled from a normal distribution with $\sigma=3.0$.
    \item By adding the encoders' power to the total cost, this gain is reduced to 14\%.
    \item The total area of the GEMM operator increases by 5\%.
\end{enumerate}

In fact, one could envision storing model weights directly in the custom encoding to further reduce system-wide latency and energy consumption.
% One could even imagine storing the weights of networks directly in the custom encoding, to reduce latency and energy consumption of the overall system.

% \begin{table}[h]
%     \centering
%     \caption{FLOWY Higher Effort Synthesis of Encoders ($\sigma=3$)}
%     \begin{tabular}{|c|c|c|c|c|}
%     \multirow{2}{*}{Metric}&\multicolumn{4}{c|}{LUT}\\
        
%             &TC&SME&GR13&GR14\\
%             \hline

%     \end{tabular}
%     % \vspace{-8pt}
%     \label{tab:high_effort_synth_encoders}
% \end{table}

\section{Limitations}

While GENIAL demonstrates that encoding-based DSE can uncover novel arithmetic designs, several limitations remain:

\subsubsection{Runtime of high-fidelity flows}
Our framework offers multiple flows with different fidelity/cost trade-offs.
Complexity (\textit{CPX}) and Yosys-based transistor count (\textit{NT}) can be extracted very quickly, while high-effort synthesis optimization (FLOWY) provides more realistic estimates but requires significant runtime.
This limits the size of the search space that can be explored at full fidelity.
In practice, we balance speed and accuracy by mixing flows depending on the experiment, but scaling to larger multipliers will require further optimization (e.g., parallelization, lightweight proxies, or adaptive flow selection).

\subsubsection{LUT-based design representation}
All candidate multipliers are generated as LUT circuits, rather than heuristics-based or manually designed RTL implementations.
This abstraction was necessary to systematically explore arbitrary input encodings while ensuring that the output remained in two’s complement, enabling compatibility with existing systems.
However, LUT-based RTL can yield suboptimal intermediate results (e.g., higher switching activity compared to hand-tuned RTL) before synthesis optimization.
Future work could investigate direct RTL templates or hybrid representations to close this gap.

\subsubsection{Scale of benchmarks}
Due to the runtime and design space explosion, our experiments focus on 4-bit multipliers and randomly generated FSMs.
Although there are no standard benchmarks for encoding-driven arithmetic design, further validation on larger and more diverse circuits, such as 8- or 16-bit multipliers, square-root units, and other arithmetic operators, would better demonstrate the scalability and generality of our approach.

% 4) System-level overhead.
% We briefly evaluated the cost of encoding conversion (Table III), but a more detailed system-level analysis is still missing. In particular, the impact of operand reuse rates, conversion latency, and power overhead in full accelerator pipelines should be investigated to confirm end-to-end efficiency.

\subsubsection{Exploration capability of the design recommendation}
While GENIAL discovers encodings that reduce switching activity, the current exploration did not yet reach or surpass the \gls{sme} format.
Enhancing the search capability of the design recommendation through improved diversity control or refined objectives may help uncover superior designs.
Additionally, considering alternative constraints or scenarios (e.g., multi-modal data distributions) could reveal new encodings that outperform known formats.

\subsubsection{Single-objective exploration guidance}
The current pipeline relies on a single proxy metric to provide an optimization signal through a surrogate model.
Future extensions could incorporate multiple proxies, or even an uncertainty-aware exploration strategy, enabling active sampling in regions of high predictive uncertainty.
This could reveal better local minima, improve search guidance, and enhance robustness across design spaces.

Together, these limitations highlight avenues for future work, including improving runtime efficiency, exploring richer and more direct RTL representations, enhancing the diversity and multi-objective guidance of the search, and scaling evaluations to larger, more complex arithmetic benchmarks.

\section{Conclusion}
\label{sec:conclusion}
% Content typically includes:

% Summary of the main findings (reiterated in light of the results)
% Interpretation of the implications
% Limitations of the study
% Suggestions for future work
% Final takeaway message

% This work presents GENIAL, a novel framework for the automatic generation and optimization of common logic circuits guided by machine learning. By integrating a tailored high-effort synthesis flow with inversion of an attention-based surrogate model, GENIAL enables efficient exploration of the design space and direct optimization of hardware metrics such as power and area.

% Our results demonstrate that GENIAL is capable of finding operand encodings for mulipliers from scratch that match area efficiency of conventional representations, such as two's complement, and discover a pareto front which extends beyond known formats.

% Design exploration is to some extend limited by the available compute budget as the \arclong(qor) extraction can take considerable effort.

% The method was applied to other logic functions with state machines and can be directly applied to any arithmetic function (such as adders) or other logic with flexible data representations. Finally it would be interesing to see what other optimizatio problems this method can be applied to.

% With this work we have empirically proven that TC and SME are are and power- efficient encodings by finidng similar formats from scratch without any knowlade of common formats.
% Furthermore, we show that the parto front between these points can extending we newly found representations.

We introduced GENIAL, a machine-learning-driven framework for the automatic generation and optimization of combinational logic circuits.
By combining a high-effort logic synthesis optimization flow with model inversion of a Transformer-based surrogate model for encoding generation, GENIAL enables sample-efficient design space exploration for optimizing key hardware metrics.
Remarkably, GENIAL rediscovers the two’s-complement encoding and identifies close variants of the sign-magnitude format from scratch, without prior knowledge, while uncovering novel representations along the area–power Pareto frontier. Notably, GR13 is a 4-bit integer encoding resulting in 18\% lower power consumption than the two's complement.
These findings validate surrogate-model inversion as a powerful strategy for circuit optimization.
Future work will explore scaling GENIAL to larger circuits, other input value distributions, and circuits not as thoroughly investigated as multipliers, unlocking broader opportunities for logic function optimization.

% \newpage

% Our experimental evaluation shows that, when applied to operand encodings for 4b multipliers, GENIAL discover a pareto front which extends beyond known formats.
% Moreover we show significant improvements on \acrlong{fsm}s, hinting at the extended generalizability of GENIAL.

% We note that the depth of design exploration is constrained by our compute budget, since quality-of-result extraction via \arclong{qor} can be computationally intensive.
% We expect it to generalize readily to other arithmetic operators (such as adders) or any logic circuit with flexible data representations.

\bibliographystyle{plain}
\bibliography{biblio/ASPDAC}

% \vspace{12pt}
% \color{red}
% IEEE conference templates contain guidance text for composing and formatting conference papers. Please ensure that all template text is removed from your conference paper prior to submission to the conference. Failure to remove the template text from your paper may result in your paper not being published.

\end{document}